\documentclass[10pt,letterpaper]{article}

\usepackage{cogsci}

\cogscifinalcopy 

\usepackage{pslatex}
\usepackage{apacite}
\usepackage{float} 

\usepackage{tikz-dependency}
\usepackage{verbatim}
\usepackage{amsmath}
\usepackage{epsfig}
\usepackage{multirow}
\usepackage{linguex}
\usepackage{graphicx}
\usepackage{subfig}
\usepackage{devanagari}
\usepackage{wrapfig}
\usepackage{cases}
\usepackage{color}
\usepackage{booktabs}
\usepackage{nameref}
\usepackage[titletoc]{appendix}
\usepackage{mathtools}
\usepackage{lipsum}  
\usepackage{comment}
\usepackage{xurl}
\usepackage{commath}
\usepackage{gb4e}

\title{Work Smarter...Not Harder!\\ Efficient Minimization of Dependency Length in SOV Languages}
 
\author{{\large \bf Sidharth Ranjan (sidharth.ranjan@ling.uni-stuttgart.de)} \\
 University of Stuttgart, 70174 Stuttgart, Germany  \\
  \AND {\large \bf Titus von der Malsburg (titus.von-der-malsburg@ling.uni-stuttgart.de)} \\
  University of Stuttgart, 70174 Stuttgart, Germany}

\begin{document}

\maketitle

\begin{abstract}

    Dependency length minimization is a universally observed quantitative property of natural languages. However, the extent of dependency length minimization, and the cognitive mechanisms through which the language processor achieves this minimization remain unclear. This research offers mechanistic insights by postulating that moving a short preverbal constituent next to the main verb explains preverbal constituent ordering decisions better than global minimization of dependency length in SOV languages. This approach constitutes a least-effort strategy because it’s just one operation but simultaneously reduces the length of all preverbal dependencies linked to the main verb. We corroborate this strategy using large-scale corpus evidence across all seven SOV languages that are prominently represented in the Universal Dependency Treebank. These findings align with the concept of bounded rationality, where decision-making is influenced by `quick-yet-economical' heuristics rather than exhaustive searches for optimal solutions. Overall, this work sheds light on the role of bounded rationality in linguistic decision-making and language evolution.

\end{abstract}

\section{Introduction}\label{sect:intro}

\begin{figure}
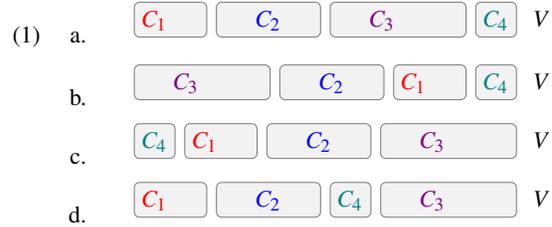

\begin{small}
\begin{exe}

  \ex \label{ex:hindi-intro}
  \begin{xlist}
  \ex[]{\label{ex:h1}
\begin{dependency}[arc edge, arc angle=80, text only label, label style={above, scale=1.8}, edge style= {black, dotted}]
\begin{deptext}[column sep=0.2cm]
{\bf \color{red}$C_1$} \& ~ \& ~ \& {\bf \color{blue}$C_2$} \& ~ \& ~ \& {\bf \color{violet}$C_3$} \& ~ \& ~ \& {\bf \color{teal}$C_4$} \& $V$ \\
\end{deptext}
\wordgroup[group style={fill=gray!10, draw=gray, inner sep=.2ex}]{1}{1}{2}{subj}
\wordgroup[group style={fill=gray!10, draw=gray, inner sep=.2ex}]{1}{3}{5}{obj1}
\wordgroup[group style={fill=gray!10, draw=gray, inner sep=.2ex}]{1}{6}{9}{pp1}
\wordgroup[group style={fill=gray!10, draw=gray, inner sep=.2ex}]{1}{10}{10}{verb}
\end{dependency}}

  \ex[] {\label{ex:h2} \begin{dependency}
\begin{deptext}[column sep=0.2cm]
~ \& {\bf \color{violet}$C_3$} \& ~ \& ~ \& ~ \& {\bf \color{blue}$C_2$} \& ~ \& {\bf \color{red}$C_1$} \& ~ \& {\bf \color{teal}$C_4$} \& $V$ \\
\end{deptext}
\wordgroup[group style={fill=gray!10, draw=gray, inner sep=.2ex}]{1}{1}{4}{subj}
\wordgroup[group style={fill=gray!10, draw=gray, inner sep=.2ex}]{1}{5}{7}{obj1}
\wordgroup[group style={fill=gray!10, draw=gray, inner sep=.2ex}]{1}{8}{9}{pp1}
\wordgroup[group style={fill=gray!10, draw=gray, inner sep=.2ex}]{1}{10}{10}{verb}
\end{dependency}}

  \ex[] {\label{ex:h3} \begin{dependency}
\begin{deptext}[column sep=0.2cm]
{\bf \color{teal}$C_4$} \& {\bf \color{red}$C_1$} \& ~ \& ~ \& {\bf \color{blue}$C_2$} \& ~ \& ~ \& {\bf \color{violet}$C_3$} \& ~ \& ~ \& $V$ \\
\end{deptext}
\wordgroup[group style={fill=gray!10, draw=gray, inner sep=.2ex}]{1}{1}{1}{subj}
\wordgroup[group style={fill=gray!10, draw=gray, inner sep=.2ex}]{1}{2}{3}{obj1}
\wordgroup[group style={fill=gray!10, draw=gray, inner sep=.2ex}]{1}{4}{6}{pp1}
\wordgroup[group style={fill=gray!10, draw=gray, inner sep=.2ex}]{1}{7}{10}{verb}
\end{dependency}}

  \ex[] {\label{ex:h4}  \begin{dependency}
\begin{deptext}[column sep=0.2cm]
{\bf \color{red}$C_1$} \& ~ \& ~ \& {\bf \color{blue}$C_2$} \& ~ \& {\bf \color{teal}$C_4$} \& ~ \& {\bf \color{violet}$C_3$} \& ~ \& ~ \& $V$ \\
\end{deptext}
\wordgroup[group style={fill=gray!10, draw=gray, inner sep=.2ex}]{1}{1}{2}{subj}
\wordgroup[group style={fill=gray!10, draw=gray, inner sep=.2ex}]{1}{3}{5}{obj1}
\wordgroup[group style={fill=gray!10, draw=gray, inner sep=.2ex}]{1}{6}{6}{pp1}
\wordgroup[group style={fill=gray!10, draw=gray, inner sep=.2ex}]{1}{7}{10}{verb}
\end{dependency}}

  \end{xlist}

\end{exe}
\end{small}
\vspace{-1em}
\caption{Preverbal constituent ordering problem in a language with \textsc{subject-object-verb} (SOV) word order; Constituent (C$_{i}$) is defined as a word or a group of words that functions as a single unit within a syntactic configuration}
\label{fig:word-order-problem}
\end{figure}

The syntax of natural language enables humans to achieve what \citeA{Humboldt1836} and \citeA{chomsky1965aspects} famously described as the `infinite use of finite means.' This phenomenon allows for the creation of a diverse range of expressions, such as sentences and discourses, by combining a finite set of linguistic units, namely words and constituents or phrases. However, among the myriad sequences available to convey an idea, the speaker produces just one (refer to Figures \ref{fig:word-order-problem} and \ref{fig:ordering-strategies} for an illustration). What factors influenced this decision, and what cognitive mechanisms underlie this decision-making process?

Dependency Locality Theory~(DLT; \citeNP{gibson00,Gib98}) has been very influential in predicting such ordering decisions in natural languages, based on the idea that the human mind grapples with limited working memory capacity~\cite{Liu2008,GildeaT10,futrell2015,futrell2020dependency,cog:raja,Liu2017,temperley-gildea-ar18,cog:sid}. DLT predicts that language processing system strives to maintain syntactically related words (head-dependent pair) in close proximity within the sentence in order to minimize memory load. In SVO languages, the placement of \textsc{short-before-long} constituents and in SOV languages, \textsc{long-before-short} constituents in the sentence have consistently been identified as the most preferred syntactic ordering choices, as they minimize overall dependency length of the sentences~\cite{hawkins1994,hawkins04,yamashitaChang2001,Temperley2007,RanjanMalsburg2023CogSci,zafar2023dependency}.  However, the extent to which dependency length minimization is employed in a given language is not well understood, and the cognitive mechanisms through which this minimization is accomplished also remain unclear.

This study lays the foundation for a mechanistic account that explains dependency length minimization (DLM) for preverbal constituent ordering preferences in SOV languages, including {\it Basque, Hindi, Japanese, Korean, Latin, Persian, and Turkish}. We test the hypothesis that language users minimize a sentence’s dependency length by placing only a short preverbal constituent (the shortest if possible) next to the main verb, and we posit that this better explains preverbal constituent ordering in SOV languages compared to global minimization of dependency length. We deploy large-scale corpus analyses and computational simulations using data from the Universal Dependency Treebank~(UD; \citeNP{ud-treebank-2.11}) to investigate our hypothesis. We refer to this hypothesis as a `least-effort' strategy because it concurrently shortens the length of all preverbal dependencies connected to the main-verb without the need to simulate through an entire search space of possible constituent orders for globally minimizing the dependency length within the sentence.


Our hypothesis is grounded in a substantial body of evidence from behavioral economics, specifically the principles of bounded rationality \cite{Simon1955, simon1982models, simon1990invariants, simon1991cognitive, gigerenzer2002bounded, gigerenzer2011heuristic, gigerenzer2015simply}. These principles suggest that human decision-making, constrained by cognitive resource limitations, has evolved to prioritize choices that are satisfactory rather than strictly optimal. In the current work, the proposed least-effort strategy serves as a satisfactory solution for the constituent ordering task, as it streamlines the search space of possible constituent orders for efficient communication.

To test ordering preferences, we conducted a simulation in which we created various counterfactual variants by randomly permuting the preverbal constituents of sentences sourced from the UD Treebank. The preverbal domain serves as the primary locus of constituent order variation in SOV languages, with fewer post-verbal constituents~\cite{cog:sid}. Conceptually, Fig. \ref{ex:hindi-intro} illustrates this generation process for an SOV sentence containing four preverbal constituents ($C_i$) directly dependent on the main verb ($V$). The reference sentence in Example \ref{ex:h1}, as it originally appeared in the corpus, served as the basis for generating variant sentences in Examples \ref{ex:h2} to \ref{ex:h4}, among others (4! = 24 such possibilities). Sentences that originally appeared in the corpus are considered a human preferred syntactic choice over those that are counterfactually generated~\cite{ranjanSchijndel2024CogSci}. Next, we conducted corpus analysis, where we examined the distributions of length of preverbal constituents and total dependency lengths within reference and variant sentences. We then deployed these two features in a logistic regression classifier for distinguishing corpus reference sentences from the counterfactual variant sentences. 

Our results indicate that naturally produced corpus sentences largely conform to the constituent orders predicted by the hypothesized least-effort strategy (\textit{i.e.} placing only a short preverbal constituent next to the main verb) across all tested SOV languages. Furthermore, the pressure to employ this strategy increases as the number of preverbal constituents grows, potentially suggesting that speakers aim to balance their production effort with their available computational capacity. Notably, for the task of classifying corpus reference sentences amidst counterfactually generated variants, this strategy significantly improves the accuracy of the regression model above and beyond the total dependency length across SOV languages, suggesting that it represents a `satisficing' (satisfactory and sufficient) solution that humans tend to employ when making constituent ordering decisions.

Our primary contribution is that we demonstrate the role of bounded rationality in constituent ordering decisions across SOV languages and offer cross-linguistic evidence imperative for the development of the theories of language and cognitive science~\cite{norcliffe2015cross}.

\section{Data and Methods}\label{sect:method}

\begin{figure*}[t]
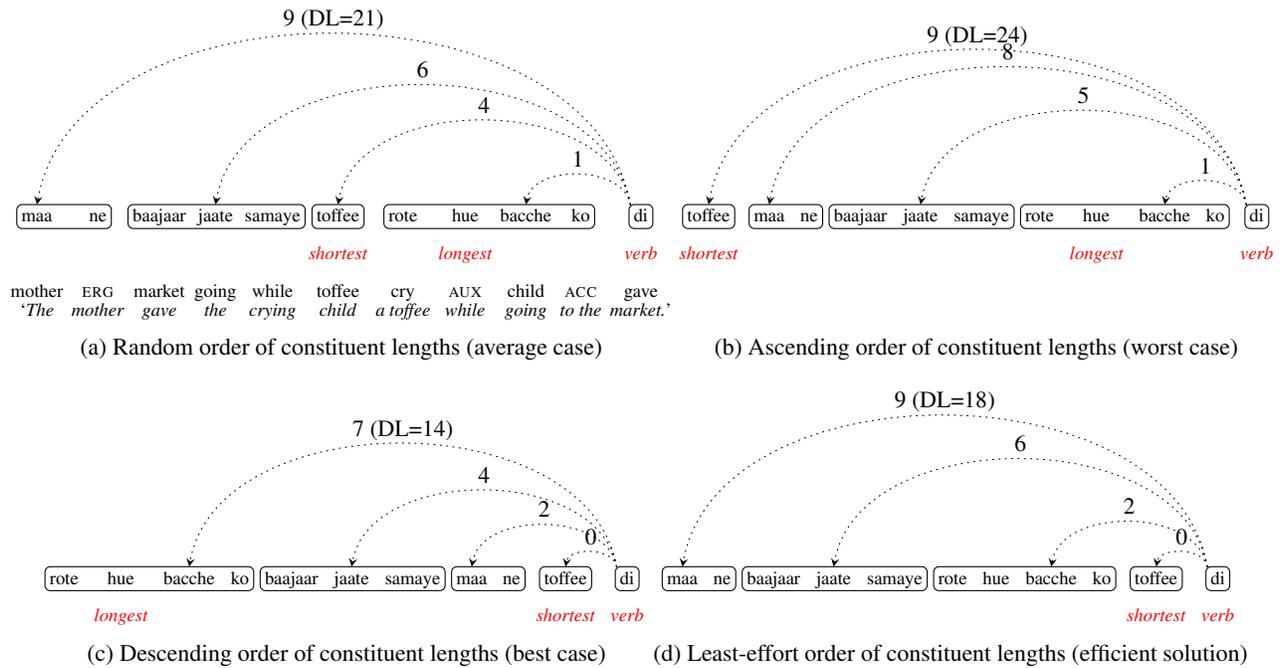

\noindent\makebox[\textwidth]{%
\subfloat[Subfigure 1 list of tables text][Random order of constituent lengths (average case)]{
\label{ex:random-order}
\begin{scriptsize}
\begin{dependency}[arc edge, arc angle=80, text only label, label style={above, scale=1.8}, edge style= {black, dotted}]
\begin{deptext}[column sep=0.0cm]
maa \& ne \& baajaar \& jaate \& samaye \& toffee \& rote \& hue \& bacche \& ko \& di\\
\& \& \& \& \& \& \& \& \& \& \\
\& \& \& \& \& \textcolor{red}{\it shortest} \& \& \textcolor{red}{\it longest} \& \& \& \textcolor{red}{\it verb}\\
\& \& \& \& \& \& \& \& \& \& \\
mother \& \textsc{erg}  \& market \& going \& while \& toffee \& cry \& \textsc{aux} \& child \& \textsc{acc} \& gave\\
`\textit{The} \& \textit{mother} \& \textit{gave} \& \textit{the} \&  \textit{crying} \& \textit{child} \& \textit{a toffee} \& \textit{while} \& \textit{going} \& \textit{to the} \& \textit{market.}'\\
\end{deptext}
\depedge{11}{1}{9 (DL=21)}
\depedge{11}{4}{6}
\depedge{11}{6}{4}
\depedge{11}{9}{1}
\wordgroup{1}{1}{2}{subj}
\wordgroup{1}{3}{5}{obj1}
\wordgroup{1}{6}{6}{pp1}
\wordgroup{1}{7}{10}{verb}
\wordgroup{1}{11}{11}{verb}
\end{dependency}
\end{scriptsize}
}
\hspace{-1.6em}
\subfloat[Subfigure 3 list of tables text][Ascending order of constituent lengths (worst case)]{
\label{ex:ascending-order}
\begin{scriptsize}
\begin{dependency}[arc edge, arc angle=80, text only label, label style={above, scale=1.8}, edge style= {black, dotted}]
\begin{deptext}[column sep=0.1cm]
toffee \& maa \& ne \& baajaar \& jaate \& samaye \& rote \& hue \& bacche \& ko \& di\\
\& \& \& \& \& \& \& \& \& \& \\
\textcolor{red}{\it shortest}\& \& \& \& \& \& \& \textcolor{red}{\it longest} \& \& \& \textcolor{red}{\it verb}\\
\& \& \& \& \& \& \& \& \& \& \\
\& \& \& \& \& \& \& \& \& \& \\
\& \& \& \& \& \& \& \& \& \& \\
\end{deptext}
\depedge{11}{1}{9 (DL=24)}
\depedge{11}{2}{8}
\depedge{11}{5}{5}
\depedge{11}{9}{1}
\wordgroup{1}{1}{1}{subj}
\wordgroup{1}{2}{3}{obj1}
\wordgroup{1}{4}{6}{pp1}
\wordgroup{1}{7}{10}{verb}
\wordgroup{1}{11}{11}{verb}
\end{dependency}
\end{scriptsize}
}
}
\\
\noindent\makebox[\textwidth]{%
\subfloat[Subfigure 4 list of tables text][Descending order of constituent lengths (best case)]{
\label{ex:descending-order}
\begin{scriptsize}

\begin{dependency}[arc edge, arc angle=80, text only label, label style={above, scale=1.8}, edge style= {black, dotted}]
\begin{deptext}[column sep=0.1cm]
rote \& hue \& bacche \& ko \& baajaar \& jaate \& samaye \& maa \& ne \& toffee \& di\\
\& \& \& \& \& \& \& \& \& \& \\
\& \textcolor{red}{\it longest} \& \& \& \& \& \& \& \& \textcolor{red}{\it shortest} \& \textcolor{red}{\it verb}\\
\end{deptext}
\depedge{11}{3}{7 (DL=14)}
\depedge{11}{6}{4}
\depedge{11}{8}{2}
\depedge{11}{10}{0}
\wordgroup{1}{1}{4}{subj}
\wordgroup{1}{5}{7}{obj1}
\wordgroup{1}{8}{9}{pp1}
\wordgroup{1}{10}{10}{verb}
\wordgroup{1}{11}{11}{verb}
\end{dependency}
\end{scriptsize}
}
\hspace{-1em}
\subfloat[Subfigure 2 list of tables text][Least-effort order of constituent lengths (efficient solution)]{
\label{ex:les-order-A}
\begin{scriptsize}
\begin{dependency}[arc edge, arc angle=80, text only label, label style={above, scale=1.8}, edge style= {black, dotted}]
\begin{deptext}[column sep=0.1cm]
maa \& ne \& baajaar \& jaate \& samaye \& rote \& hue \& bacche \& ko \& toffee \& di\\
\& \& \& \& \& \& \& \& \& \& \\
\& \& \& \& \& \& \& \& \& \textcolor{red}{\it shortest} \& \textcolor{red}{\it verb}\\
\end{deptext}
\depedge{11}{1}{9 (DL=18)}
\depedge{11}{4}{6}
\depedge{11}{8}{2}
\depedge{11}{10}{0}
\wordgroup{1}{1}{2}{subj}
\wordgroup{1}{3}{5}{obj1}
\wordgroup{1}{6}{9}{pp1}
\wordgroup{1}{10}{10}{verb}
\wordgroup{1}{11}{11}{verb}
\end{dependency}
\end{scriptsize}
}
}
\vspace{-0.5em}
\caption{Preverbal constituent ordering in Hindi (SOV); Only main verb dependencies are depicted; Total dependency length (DL) and Constituent's DL are indicated above each arc; Image reproduced from \citeA{RanjanMalsburg2023CogSci}}
\label{fig:ordering-strategies}
\end{figure*}

Our dataset comprises sentences from all head-final (SOV) languages that have a significant amount of data available in the publicly accessible Universal Dependency Treebank \cite{ud-treebank-2.11} corpus.\footnote{Version 2.11; \small{\url{http://hdl.handle.net/11234/1-4923}}} We specifically selected SOV languages that met the criteria of having at least 2000 sentences and projective dependency trees featuring a minimum of two preverbal constituents. This selection criterion resulted in 7 languages in our dataset: \textit{Basque, Hindi, Japanese, Korean, Latin, Persian, and Turkish}.  
We implemented counterfactual variant generation for each reference sentence in our dataset. This process involved random permutation of preverbal constituents within the dependency tree, specifically those directly dependent on the root verb (see Fig. \ref{ex:hindi-intro} and Fig.~\ref{fig:ordering-strategies}). We restricted the variant generation to a maximum of 120 variants per corpus reference sentence, an arbitrary cutoff to keep our computation tractable.\footnote{Our findings are consistent regardless of chosen cutoff.} Table \ref{tab:ref-var-data} presents the total number of generated variants for each reference sentence in the corpus across the languages in our dataset.

Subsequently, we conducted quantitative analyses by examining the distributions of constituent lengths and dependency lengths within both reference and variant sentences containing varying preverbal constituents.\footnote{We conducted a similar analysis using only grammatical variants and found consistent results.} 
We define \textsc{constituent length} as the total number of words in a constituent as indicated by boxes around the words in Fig.~\ref{fig:ordering-strategies}. 
We define \textsc{dependency length} as the count of intervening words between head and dependent units within a dependency graph~\cite{Temperley2007}. Total dependency length is then calculated by summing the dependency length of each word within the sentence.
Afterwards, we set up a classification task to investigate whether the aforementioned features are effective in identifying the corpus reference sentences amidst competing counterfactual variants (see Table \ref{tab:ref-var-data}). For corpus analyses reported in this paper, 
we focus on sentences with five or fewer preverbal constituents due to data sparsity issues with those containing six or more.
However, we use the entire dataset, including sentences with 6 or more preverbal constituents, to compute our prediction results.

\begin{table}
    \scalebox{0.65}{
    \begin{tabular}{l|ccccccc}
        \textbf{Type}  & \textbf{Hindi}  & \textbf{Japanese} & \textbf{Korean} & \textbf{Turkish} & \textbf{Basque} & \textbf{Persian} & \textbf{Latin}  \\\hline
        \textbf{Reference} & ~~10,223  & ~~24,198    & ~~18,868  & ~~25,928   & ~~4,628   & ~~23,969   & ~~17,294  \\
        \textbf{Variant}   & 308,571 & 325,958   & 262,696 & 204,783  & 35,988  & 618,070  & 191,053 \\\hline
    \end{tabular}}
    \caption{Total number of reference sentences that originally appeared in the corpus and corresponding artificially generated variant sentences in our dataset}
    \label{tab:ref-var-data}
\end{table}
\begin{figure}[t]
  \centering
  \includegraphics[scale=0.27]{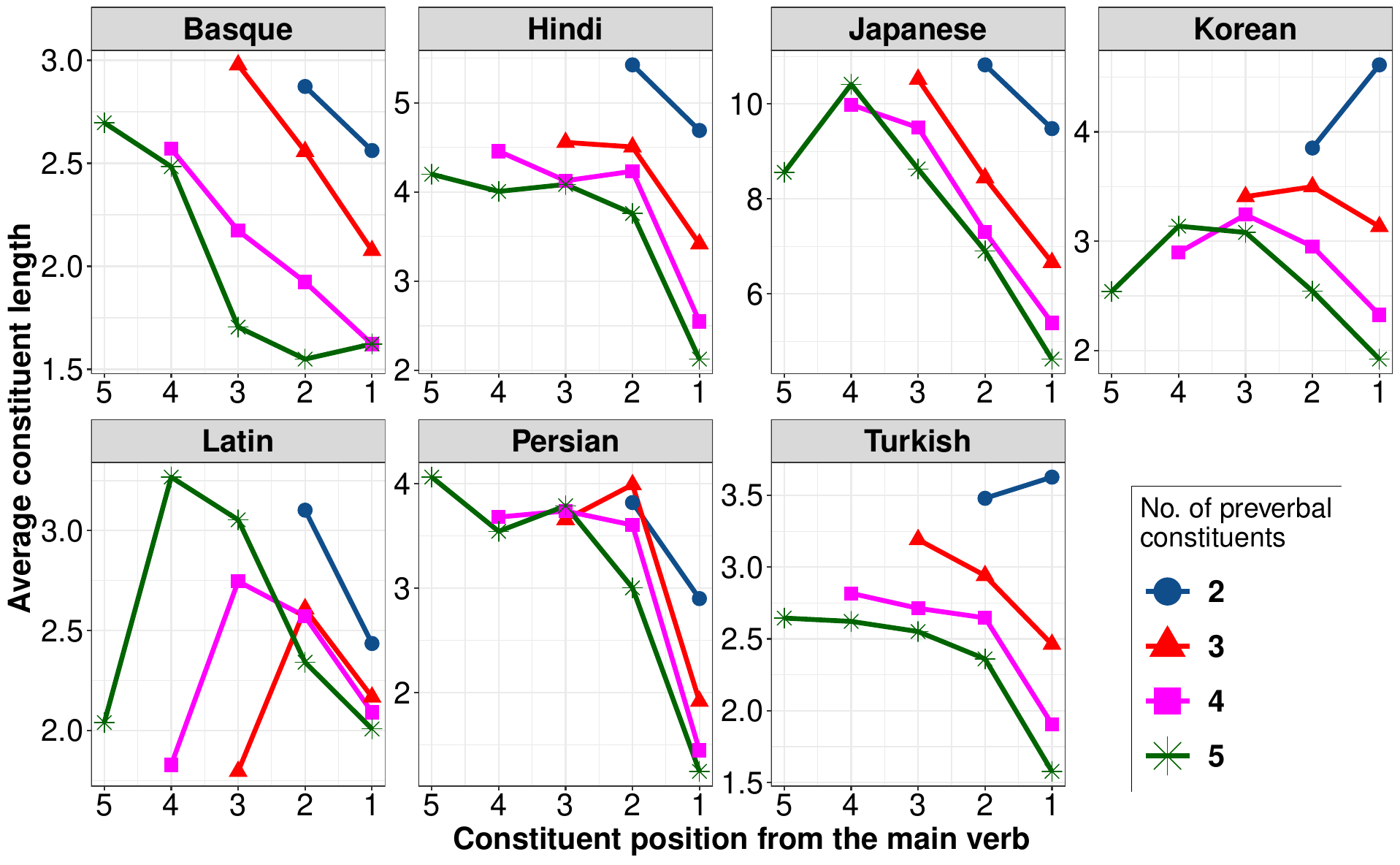}
  \caption{Average constituent length of preverbal constituents
    for corpus sentences with 2 to 5 constituents separately}
  \label{fig:constlen-dist-lang}
\end{figure}

\section{Results}\label{sect:exp}

\subsection{Constituent Length Analysis}\label{sect:constlen-analysis}

According to our least-effort strategy, the preverbal constituent adjacent to the main verb should be amongst the short preverbal constituents in the sentence. In contrast, the global DLM would predict a gradual decrease in the lengths of preverbal constituents as they approach the main-verb. To test these predictions, we computed the average length of constituents at various preverbal positions in the sentences from the original corpus. Fig.~\ref{fig:constlen-dist-lang} displays the average constituent length distribution for corpus sentences with 2 to 5 preverbal constituents in SOV languages. 

In Persian, the average constituent length remains relatively consistent across positions until the position next to the main verb, where the length suddenly decreases. This suggests that the constituent immediately adjacent to the main verb is generally shorter than constituents in any other preverbal position in the sentence. This trend is consistent across all sentences types with varying numbers of preverbal constituents, confirming the least-effort strategy in Persian. A similar pattern is observed in Hindi. Turkish and Korean also follow a similar pattern for the most part, but with a deviation observed in sentences with two preverbal constituents. This deviation may indicate that these languages do not prioritize optimizing dependency length due to the low memory pressure associated with these sentences. Additionally, in Korean, while the shortest constituent on average is positioned next to the main verb, corpus sentences also often begin with a short constituent.
These short constituents generally include left dislocated elements (29\%) akin to clefting~\cite{fernandez2020dislocations}, conjunctions (14\%), clausal complements (13\%), \textit{inter alia}. This suggest that certain grammatical constraints, in addition to factors like information structure, discourse, and style, may take precedence over processing considerations. The plots for Basque and Japanese exhibit a more gradual decrease in average constituent length as one moves towards the main verb, with the on-average shortest constituent positioned next to the main verb. 

Interestingly, in Latin (akin to patterns in Korean), while the shortest preverbal constituent is commonly positioned in the first position, the constituent at the last preverbal position is also short. Further linguistic analysis of Latin sentences revealed a specific stylistic preference, where a lot of sentences begin with the single-word conjunctions (23\%) like “et”,\footnote{Example (Thomas Acquinas): \textit{Et ita deum esse per se notum erit.} (`And so it will be known that god exists by himself.') In terms of dependency length, \textit{et} is an ideal candidate for placement next to the verb, but grammar prevents it.} in addition to other categories (22\%) involving short phrases such as adverbial modifiers and negations, thereby considerably reducing the size of the first constituent in the sentence. 
Despite this stylistic preference in the corpus, the pressure to maintain a short preverbal constituent adjacent to the main verb compellingly illustrates the existence of expected pattern in Latin, consistent with other SOV languages under study. Additionally, Fig.~\ref{fig:constlen-dist-lang} further illustrates that the pressure to place a short preverbal constituent adjacent to the main verb across SOV languages increases as the number of preverbal constituents with sentences grows, possibly due to increased memory load.

Taken together, these plots offer crucial evidence that naturally occurring sentences consistently exhibit a preference for optimizing only or at least primarily the length of the constituent next to the main verb in SOV languages. This approach may be employed to achieve a balance between production effort and the cognitive constraints experienced by listeners consistent with core assumptions of the framework of bounded rationality.

\subsection{Counterfactual Analysis}\label{sect:countfact-analysis}

In this analysis, our objective is to simulate human preferences for constituent orders across SOV languages. We aim to identify which constituent orders closely align with the dependency lengths observed in the corpus data, a true representative of human behavior. For this, we compared the total dependency length of corpus reference sentences to four different types of variants, as shown in Fig.~\ref{fig:ordering-strategies}.
\smallskip

\begin{small}
    \begin{enumerate}
  \item \textsc{Random order:} Randomly arrange the preverbal constituents in the sentence (average case, Fig.~\ref{ex:random-order}).
  \item \textsc{Ascending order:} Arrange the preverbal constituents in increasing order of constituent lengths, resulting in maximal dependency length of the sentence (worst case, Fig.~\ref{ex:ascending-order}).
  \item \textsc{Descending order:} Arrange the preverbal constituents in decreasing order of constituent lengths, globally minimizing the dependency length of the sentence (best case, Fig.~\ref{ex:descending-order}).
  \item \textsc{least-effort order:} Start with any random order of preverbal constituents and then move shortest constituent next to main verb (efficient solution, Fig.~\ref{ex:les-order-A}).
      \end{enumerate}
\end{small}

\begin{figure}[!ht]
  \centering
  \includegraphics[scale=0.27]{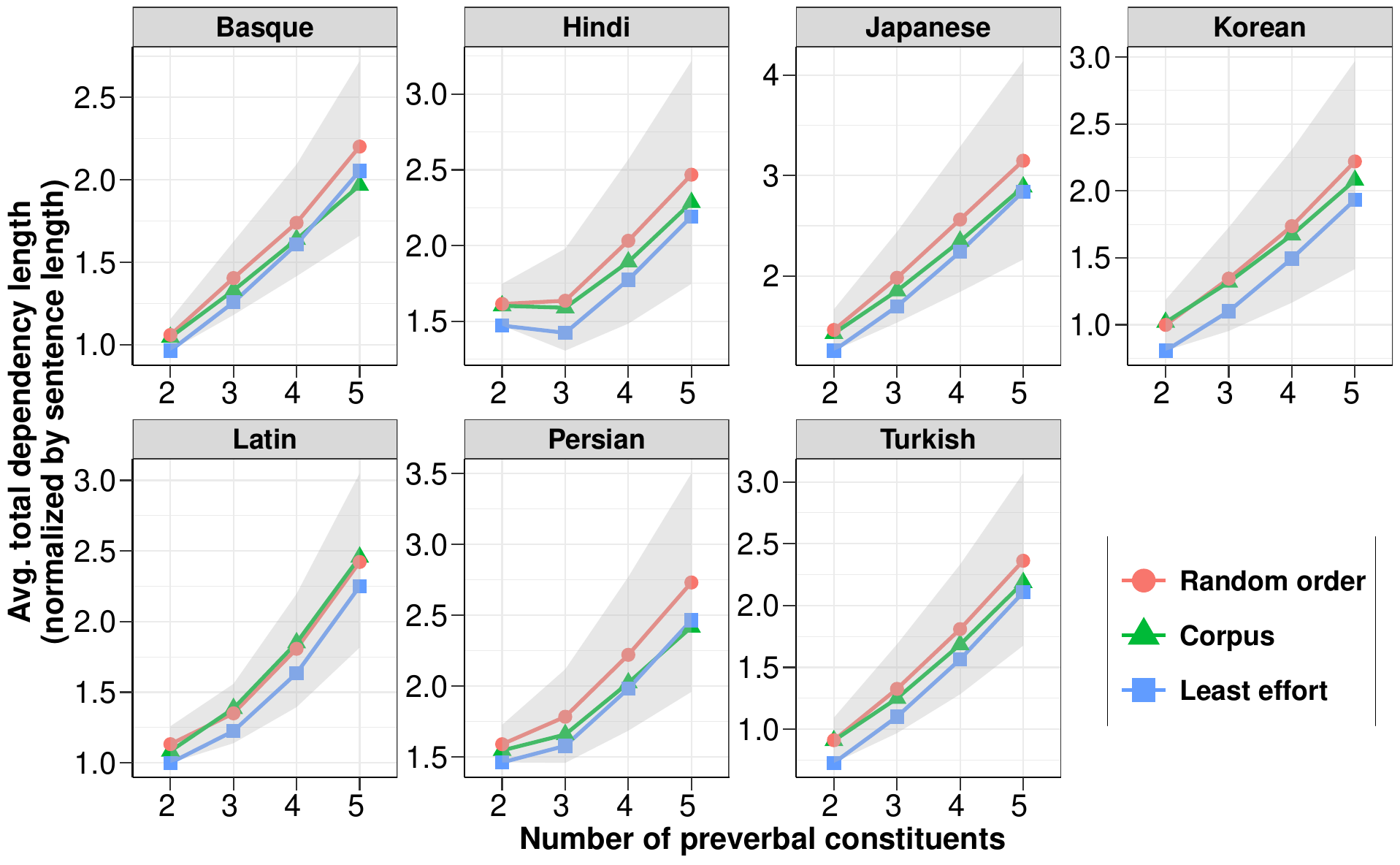}
  \caption{Average total dependency length (\textit{i.e.,} total dependency length normalized by the number of words in a sentence) for different constituent orderings and for different numbers of preverbal constituents}
  \label{fig:deplen-dist-lang}
\end{figure}

Fig.~\ref{fig:deplen-dist-lang} illustrates the trends in average dependency lengths across reference and variant sentences with varying preverbal constituents and ordering types. The grey shaded region in the plot represents the entire spectrum of dependency length of sentences normalized by their sentence lengths, ranging from the maximal dependency length (achieved via ascending order of preverbal constituent lengths) to the minimal dependency length (via descending order). This shaded region serves as the strict boundary within which the language processor can adjust its dependency length by reordering preverbal constituents. Interestingly, we observe that the dependency length of corpus sentences, across languages, aligns with the dependency length of sentences predicted by least-effort. In essence, the dependency length obtained by the least-effort strategy estimates the lower bound to the corpus sentences, indicating that the language system typically does not minimize dependency length beyond this point.

Through this simulation we enforced a strict constraint on the hypothesized least-effort strategy: first, identify the shortest preverbal constituent, and then position it next to the main verb. A boundedly rational language system would efficiently operate within this lower bound by placing only a short preverbal constituent next to the main verb, without concerning itself with other preverbal constituents. This strategic optimization adopted by the speakers aims to minimize the sentence's dependency length, reflecting considerations of memory load.
Consistent with these insights, Fig.~\ref{fig:deplen-dist-lang} also suggests that the tendency for corpus sentences to align with the least-effort solution strengthens as the number of preverbal constituents---and therefore memory load---goes up. 

Additionally, based on these plots, we also find that corpus sentences, on average, exhibit lower dependency length than sentences with random preverbal constituent orderings. This trend aligns with previous studies~\cite{Liu2008, GildeaT10, futrell2015}, 
except for Latin and, perhaps, Korean. As previously discussed, a significant number of corpus sentences in Latin begin with a single-word categories directly linked to the main verb. In case of Korean, although the shortest constituent is often next to the main verb, the sentences generally start with a short constituent. This results in a substantially longer dependency length within sentences for both the languages. Nonetheless, we contend that our least-effort strategy still holds in both Latin and Korean (see Fig. \ref{fig:constlen-dist-lang}), as there is a consistent effort to reduce the preverbal constituent length next to the main verb to the extent possible. Therefore, we conjecture that a greater average dependency length for corpus sentences compared to sentences with random orders does not necessarily imply the absence of dependency length minimization in a language~\cite{ferrer2014risks}. In summary, these observations reinforce the efficacy of the least-effort strategy in predicting constituent ordering choices in SOV languages. 

\subsection{Corpus Prediction Task}\label{sect:ref-pred-task}

If speakers employ the hypothesized least-effort strategy, we should be able to predict whether a sentence is a corpus reference sentence or a counterfactual variant by only examining the length of the preverbal constituent next to the main verb. Further, these predictions should be \textit{better} than those obtained when total dependency length is used as the predictor. For this, we deployed logistic regression model to identify reference sentences amidst the generated variants using features such as preverbal constituent length next to the main verb (CL Last) and/or total dependency length (Total DL).
Table \ref{tab:ref-var-data} presents the count of corpus reference sentences and alternative variants for each language in our dataset. The presence of a greater number of variants per reference sentence introduces a substantial class imbalance issue (\textsc{reference} / \textsc{variant}) for our binary classification task. The subsequent section addresses this class imbalance problem by transforming the dataset (explained below) with balanced class labels and then reports computational modeling analyses in Table \ref{tab:comp-model}.

\subsubsection{Ranking Model}\label{sect:rank-model}

In order to address the class imbalance issue for appropriate classification, we adopt an approach proposed by \citeA{Joachims:2002}, originally designed for ranking web pages. Previous work dealing with the syntactic choice prediction task~\cite{cog:sid}, similar to our current work, has also employed this method. This approach converts a binary classification task into a pairwise ranking task involving feature vectors of a reference sentence and each of its variants. In this case, the feature vectors consist of the input features used for each reference and variant sentence in the model. Subsequently, we train a logistic regression model on the difference between feature vectors of reference and variant pairs, as illustrated in the equations below:

\begin{small}
  \begin{equation}
    \label{eq:refvar1}
    \mathbf{w}\cdot\phi(Reference)>\mathbf{w}\cdot\phi(Variant)
  \end{equation}
  \begin{equation}
    \label{eq:refvar2}
    \mathbf{w}\cdot\mathbf{(}\phi(Reference)-\phi(Variant)\mathbf{)}>0
  \end{equation}
\end{small}

\noindent
Equation~\ref{eq:refvar1} represents a standard classification model that evaluates whether the reference sentence ranks higher than one of its variants. This decision is made by comparing the dot product of the feature vector associated with the reference sentence and the learned feature weights $\mathbf{w}$ with the corresponding dot product of the variant sentence. This relationship can also be expressed in the form of Equation~\ref{eq:refvar2}, where feature values of the first member are subtracted from the second member~\cite{Joachims:2002}. Now, the model's decision for a particular referent-variant pair can be made by evaluating the sign of the dot product involving learned feature weight and difference between the feature vectors (see Equation ~\ref{eq:refvar2}).

We created ordered pairs consisting of feature vectors of reference (\textsc{ref}) and variant (\textsc{var}) sentences, ensuring that the counts of orders of each type (\textsc{ref-var}, \textsc{var-ref}) are balanced. 
For instance, the reference sentence \ref{ex:h1} in Fig. \ref{fig:word-order-problem} would generate three such pairs: (\ref{ex:h1}-\ref{ex:h2}), (\ref{ex:h3}-\ref{ex:h1}), and (\ref{ex:h1}-\ref{ex:h4}), where sentence (\ref{ex:h1}) is the corpus reference sentence since it happens to be originally present in the corpus, and rest all are counterfactual variants (\ref{ex:h2}-\ref{ex:h4}). 
The pairs that alternate between `\textsc{ref-var}' were assigned the label `1,' while the pair `\textsc{var-ref}' received the label `0.' This alternate coding results in a balanced dataset and contains an equal number of labels of each type when the total number of variants is even, and a difference of one label when the total is odd.

The transformed feature values were incorporated into a logistic regression model using the \texttt{glm} function in R. We used the following \texttt{glm} equation\footnote{We followed the R \texttt{GLM} format: the dependent variable is on the left of `$\sim$,' and the independent variables are on the right, with $\delta$ indicating feature differences.} 
to investigate our hypothesis: $choice \sim  \delta~constituent~length$ and $choice \sim  \delta~dependency~length$. Here, choice is a binary choice dependent variable (1 denotes reference sentence preference, and 0 denotes variant sentence preference). The delta ($\delta$) refers to the difference between the feature vectors of reference sentence and its paired variant. All the independent variables (CL Last and Total DL) were normalised to $z$-scores, \emph{i.e.,} the predictor’s value (centered around its mean) was divided by its standard deviation.  We evaluate our model's performance using 10-fold cross-validation and report the classification accuracy in Table \ref{tab:classification-results}. 
However, for estimating regression coefficients shown in Table~\ref{tab:regression-results}, we used the entire transformed test data for our experiments. 
The Pearson's correlation coefficient between `$\delta$ CL Last' and `$\delta$ Total DL' was consistently around 0.70 for all languages.

\subsubsection{Regression Analysis}\label{sect:reg-analysis}

\begin{table*}
\centering
\begin{small}
\subfloat[Regression coefficients of distinct models with constituent length (CL) of last preverbal constituent and total dependency length (Total DL) as predictors; all regression coefficients are significant with $p$~\textless~0.001]
{
    \begin{tabular}{l|ccc}
        Language & CL Last & Total DL & Total DL + CL Last \\\hline
         Basque &   -0.41 & -0.83 & -0.95, ~0.17 \\
         Hindi &    -1.15 & -0.73 & -0.18, -1.02 \\
         Japanese & -0.71 & -0.68 & -0.40, -0.44 \\
         Korean &   -0.37 & -0.31 & -0.12, -0.29 \\
         Latin &    -0.33 & -0.26 & -0.08, -0.28 \\
         Persian &  -2.96 & -1.22 & -0.31, -2.76 \\
         Turkish &  -0.61 & -0.55 & -0.27, -0.42 \\\hline
    \end{tabular}
    \label{tab:regression-results}
}
\qquad\quad\qquad
\subfloat[Classification accuracy (\%) of distinct models (10-fold cross-validation) with constituent length (CL) of last preverbal constituent and total dependency length (Total DL) as predictors (Random accuracy = 50\%)]
{
    \begin{tabular}{l|ccc}
        Language & CL Last & Total DL & Total DL + CL Last \\\hline
         Basque &   55.07 & 61.71 & 62.01 \\
         Hindi &    69.49 & 63.39 & 69.23 \\
         Japanese & 62.80 & 63.09 & 64.36 \\
         Korean &   56.92 & 55.11 & 56.44 \\
         Latin &    51.48 & 48.51 & 49.55 \\
         Persian &  74.57 & 69.04 & 75.17 \\
         Turkish &  61.72 & 60.00 & 62.02 \\\hline
    \end{tabular}
    \label{tab:classification-results}
}
\end{small}
\vspace{-0.7em}
\caption{Logistic regression models determining corpus \textsc{reference} sentence and its paired counterfactual \textsc{variant}}
\label{tab:comp-model}
\end{table*}

Table \ref{tab:regression-results} presents the regression coefficients (all significant with $p$~\textless~0.001) of different models on the entire test dataset for each language under investigation. In the models with individual feature (second column in Table \ref{tab:regression-results}), the negative regression coefficient for `CL Last' suggests that sentences naturally occurring in the corpus tend to have shorter constituents adjacent to the main verb than alternative variants across all SOV languages, thus supporting the presence of the least-effort strategy. Additionally, consistent with previous studies, the negative regression coefficient for `Total DL' in the model with individual feature (third column in Table \ref{tab:regression-results}) suggests that sentences in SOV languages consistently minimize their dependency length owing to memory constraints. When both these features are included in the model (see the last column in Table \ref{tab:regression-results}), we consistently observe negative regression coefficients for both features across SOV languages, except for Basque where the coefficient sign for `CL Last' flips. The variance inflation factor (VIF) for each feature in the aforementioned combined model consistently remained below 1.75 across all analyzed languages. We discuss Basque in the next subsection.

\subsubsection{Prediction Accuracy}\label{sect:class-analysis}

Table \ref{tab:classification-results} presents the classification performance of our various models on held-out dataset via 10-fold cross-validation to determine how many sentences are affected by each predictor. In terms of individual performance, the model with feature `CL Last', corresponding to the least-effort strategy (second column in Table \ref{tab:classification-results}), accounted for the majority of our data for all SOV languages, except Basque and Japanese. In these two languages, `Total DL' achieved slightly higher prediction accuracy than `CL Last', with a small but statistically significant margin (p $<$ 0.001 using McNemar's two-tailed significance test). Crucially, however, over a baseline model containing only `Total DL' measure, adding the feature `CL Last' (see the last column in Table \ref{tab:classification-results}) induced a significant increase in classification accuracy across all SOV languages, including Basque (p $<$ 0.001 using McNemar's two-tailed significance test compared to the previous column in Table \ref{tab:classification-results}). Therefore, despite the sign flip in the regression coefficient of `CL Last' in the combined model for Basque (last column in Table \ref{tab:regression-results}), the significant increase in 10-fold cross-validation accuracy suggests that the preverbal constituent next to the main verb, typically shorter in reference sentences than in the paired variants, continues to play an important role in Basque as well. Thus these observations highlight the effectiveness of our least-effort strategy in predicting constituent ordering choices across SOV languages.

\section{Discussion}\label{sect:disc}

While the minimization of dependency length has been recognized as a universal quantitative property of natural languages, this work investigates the extent of minimization and the cognitive mechanisms driving the process. We found that speakers of SOV languages, consistent with the central ideas of bounded rationality in decision-making, choose constituent orders that minimize dependency length by employing a least-effort strategy as a fundamental mechanism. This strategy involves placing a short preverbal constituent, possibly the shortest one, next to the main verb, rather than aiming for the globally optimal solution, \textit{i.e.} minimizing the dependency length entirely in the sentence.
We also found that the pressure to move a short preverbal constituent next to the main verb increases as the number of preverbal constituents grows, suggesting a trade-off between production effort and limited cognitive resources across SOV languages. Finally, for the task of identifying corpus sentences amidst the alternative variants, the feature adhering to least-effort strategy significantly improved the prediction accuracy beyond total dependency length for all SOV languages under study.

Consistent with previous work~\cite{Liu2008,GildeaT10,futrell2020dependency}, our experiments show that corpus sentences consistently exhibit lower dependency length than sentences with random preverbal constituent orderings in all SOV languages, except Latin. Despite the stylistic preference to place single-word conjunctions at the beginning of Latin sentences, the preverbal constituent adjacent to the main verb still indicates an attempt to keep dependency length small, confirming the least-effort strategy for Latin as well. Our fine-grained analysis therefore shows that a lack of total dependency length minimization (compared to random constituent orders) does \textit{not} imply that producers of Latin forgo opportunities to optimize dependency length. Moreover, the example of Latin also illustrates how language users formulate sentences not only by optimizing dependency length; their choices are simultaneously influenced by many other factors, including grammar, discourse, communicative efficiency, and style. The proposed least-effort strategy necessarily operates in concert with these other constraints.

Besides, our interpretation of bounded rationality may hold for both speakers and listeners, since both rely on shared linguistic cues to navigate the challenges associated with word order flexibility~\cite{ChaterChristiansen2016, momma2018}. 
Future work needs to test the proposed strategy for real-time production and comprehension, in addition to developing a full-fledged mechanistic account of the cognitive processes underlying DLM.

The broader implication of our findings is that flexibility in language and constituent order preferences can be understood through the framework of bounded rationality in human behavior. This behavior is limited by human's cognitive abilities, information availability, and the time pressure under which individuals must respond~\cite{Simon1955,simon1982models}. These perspectives emphasize that language, as a tool for communication, has evolved to align with the cognitive abilities of its users, making it both efficient and comprehensible. Simultaneously, these perspectives also offer valuable insights into how linguistic patterns and preferences emerge as a result of humans having to deal with their cognitive and informational limitations~\cite{gigerenzer2011heuristicsbook}.

Overall, our results provide converging evidence that preverbal constituent ordering preferences in SOV languages are shaped by the minimization of dependency length within the constraints of bounded rationality.

\newpage

\section{Acknowledgments}

We thank Daniel Hole, Judith Tonhauser, Lisa Hofmann, Samar Husain, Shravan Vasishth, and the attendees at the Linguistic Evidence (LE-2024) conference, the Sentence Processing colloquium at the University of Potsdam, and the Linguistic Theory colloquium at the University of Stuttgart for their insightful comments and feedback on this work. We would also like to express our gratitude to the anonymous reviewers of SAFAL-2023, EACL-2024, LE-2024, and CogSci-2024 for their invaluable comments and suggestions.

\bibliographystyle{apacite}

\setlength{\bibleftmargin}{.125in}
\setlength{\bibindent}{-\bibleftmargin}

\bibliography{deplen,extra}

\end{document}